\documentclass[10pt,twocolumn,letterpaper]{article}

\usepackage{iccv}
\usepackage{times}
\usepackage{epsfig}
\usepackage{graphicx}
\usepackage{amsmath}
\usepackage{amssymb}
\usepackage{url}
\usepackage{enumitem}
\usepackage{soul}
\usepackage{multirow}
\usepackage{multicol}
\usepackage{booktabs}
\usepackage{appendix}

\newcommand{\mv}[1]{\mathbf{#1}}
\newcommand{\aroundEqn}{\vspace{-0.2em}}
\usepackage[pagebackref=true,breaklinks=true,letterpaper=true,colorlinks,bookmarks=false]{hyperref}

\iccvfinalcopy %

\def\iccvPaperID{917} %

\newcommand{\bd}[1]{\textbf{#1}}
\newcommand{\ud}[1]{#1}

\usepackage{lipsum}
\newcommand\blfootnote[1]{%
  \begingroup
  \renewcommand\thefootnote{}\footnote{#1}%
  \addtocounter{footnote}{-1}%
  \endgroup
}

\DeclareMathOperator*{\argmin}{arg\,min \quad}

\ificcvfinal\fi
\begin{document}

\title{SENSE: a Shared Encoder Network for Scene-flow Estimation}

\author{
Huaizu Jiang$^{1\dag}$~~~~~~~~Deqing Sun$^{2*}$~~~~~~~~Varun Jampani$^{2*}$\\
Zhaoyang Lv$^{3\dag}$~~~~~~~~Erik Learned-Miller$^1$~~~~~~~~Jan Kautz$^2$\\
$^1$UMass Amherst~~~~~$^2$NVIDIA~~~~~$^3$Georgia Tech\\
}

\maketitle
\ificcvfinal\thispagestyle{empty}\fi

\begin{abstract}
We introduce a compact network for holistic scene flow estimation, called SENSE, which shares common encoder features among four closely-related tasks: optical flow estimation, disparity estimation from stereo, occlusion estimation, and semantic segmentation. Our key insight is that sharing features makes the network more compact, induces better feature representations, and can better exploit interactions among these tasks to handle partially labeled data. %
With a shared encoder, we can flexibly add decoders for different tasks during training. This modular design leads to a compact and efficient model at inference time. Exploiting the interactions among these tasks allows us to introduce distillation and self-supervised losses in addition to supervised losses, which can better handle partially labeled real-world data.
SENSE achieves state-of-the-art results on several optical flow benchmarks and runs as fast as networks specifically designed for optical flow. It also compares favorably against the state of the art on stereo and scene flow, while consuming much less memory.
\blfootnote{\dag The work was begun while the author was an intern at NVIDIA.}
\blfootnote{*Currently affiliated with Google.}%
\blfootnote{Code is available at: \url{https://github.com/NVlabs/SENSE}}
\end{abstract}

\section{Introduction}

Scene flow estimation aims at recovering the 3D structure (disparity) and motion of a scene from image sequences captured by two or more cameras~\cite{vedula1999three}. It generalizes the classical problems of optical flow estimation for monocular image sequences and disparity prediction for stereo image pairs.
There has been steady and impressive progress on scene flow estimation, as evidenced by results on the KITTI benchmark~\cite{menze2015object}. State-of-the-art scene flow methods outperform the best disparity (stereo) and optical flow methods by a significant margin, demonstrating the benefit of additional information in the stereo video sequences.
However, the top-performing scene flow methods~\cite{Behl2017ICCV,vogel20153d} %
are based on the energy minimization framework~\cite{Horn:1981:DO} and are thus computationally expensive for real-time applications, such as 3D motion capture~\cite{furukawa10dense} and autonomous driving~\cite{JanaiGBG17}.

\begin{figure}
\centering
\renewcommand{\tabcolsep}{0.25mm}
{\small
\begin{tabular}{ll}
\includegraphics[width=0.5\linewidth]{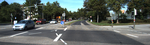} &
\includegraphics[width=0.5\linewidth]{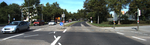} \\
{\small (a) Image 1 of left camera } & {\small (b) Image 2 of left camera }\\
\includegraphics[width=0.5\linewidth]{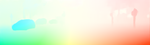} &
\includegraphics[width=0.5\linewidth]{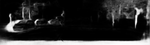} \\
(c) Optical flow  & (d) Occlusions for flow \\
\includegraphics[width=0.5\linewidth]{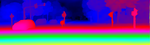} &
\includegraphics[width=0.5\linewidth]{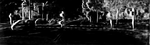} \\
(e) Disparity  & (f) Occlusions for disparity\\
\includegraphics[width=0.5\linewidth]{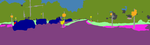} &
\includegraphics[width=0.5\linewidth]{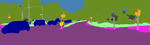} \\
(g) Segmentation of (a) & (h) Segmentation of (b)\\
\end{tabular}
}
\vspace{1mm}
\caption{Given stereo videos, we train compact networks for several holistic scene understanding problems by sharing features.
}
\label{fig:teaser}
\end{figure}

Recently, a flurry of convolutional neural network (CNN)-based methods have been developed for the sub-problems of stereo and optical flow. These methods  achieve state-of-the-art performance and run in real-time. However, while stereo and flow are closely-related, the top-performing networks for stereo and flow adopt significantly different architectures. %
Further, existing networks for scene flow stack sub-networks for stereo and optical flow together~\cite{Mayer:2016:Large,Ilg2018occlusions}, which does not fully exploit the structure of the two tightly-coupled problems.

As both stereo and flow rely on pixel features to establish correspondences, will the same features work for these two or more related tasks? To answer this question, we take a modular approach and build a Shared Encoder Network for Scene-flow Estimation (SENSE).
Specifically, we share a feature encoder among four closely-related tasks: optical flow, stereo, occlusion, and semantic segmentation. Sharing features makes the network compact and also leads to better feature representation via multi-task learning.

The interactions among closely-related tasks further constrain the network training, ameliorating %
the issue of sparse ground-truth annotations for scene flow estimation. Unlike many other vision tasks, it is inherently difficult to collect ground-truth optical flow and stereo for real-world data. Training data-hungry deep CNNs often relies on synthetic data~\cite{Butler:ECCV:2012,Dosovitskiy:2015Flownet,Mayer:2016:Large}, which lacks the fine details and diversity ubiquitous in the real world.
To narrow the domain gap, fine-tuning on real-world data is necessary, but the scarcity of annotated real-world data has been a serious bottleneck for learning CNN models for scene flow.

To address the data scarcity issue, we introduce a semi-supervised loss for SENSE by adding distillation and self-supervised loss terms to the supervised losses. First, no existing dataset provides  ground truth annotations for all the four tasks we address. For example, the KITTI benchmark has no ground truth annotations for occlusion and semantic segmentation.\footnote{Segmentation is only available for left images of KITTI 2015~\cite{Alhaija2018IJCV}.}
Thus, we train separate models for tasks with missing ground truth annotations using other annotated data, and use the pre-trained models to ``supervise'' our network on the real data via a distillation loss~\cite{hinton15distilling}. %
Second, we use self-supervision loss terms that encourage corresponding visible pixels to have similar pixel values and semantic classes, according to either optical flow or stereo. The self-supervision loss terms tightly couple the four tasks together and are critical for improvement in regions without ground truth, such as sky regions.

Experiments on both synthetic and real-world benchmark datasets demonstrate that SENSE achieves state-of-the-art results for optical flow, %
while maintaining the same run-time efficiency as specialized networks for flow.
It also compares favorably against state of the art on disparity and scene flow estimation, while having a much smaller memory footprint. Ablation studies confirm the utility of our design choices, and show that our proposed distillation and self-supervised loss terms help mitigate issues with partially labeled data.

To summarize, we make the following contributions:
\begin{itemize}[topsep=0pt,noitemsep]
    \item We introduce a modular network design for holistic scene understanding, called SENSE, to integrate optical flow, stereo, occlusion, and semantic segmentation.
    \item SENSE shares an encoder among these four tasks, which makes networks compact and also induces better feature representation via multi-task learning.
    \item SENSE can better handle partially labeled data  by exploiting interactions among tasks in a semi-supervised approach; it leads to qualitatively better results in regions without ground-truth annotations.
    \item SENSE achieves state-of-the-art flow results while running as fast as specialized flow networks. It compares favorably against state of the art on stereo and scene flow, while consuming much less memory.
\end{itemize}

\section{Related Work}
A comprehensive survey of holistic scene understanding is beyond our scope and we review the most relevant work.

\noindent\textbf{Energy minimization for scene flow estimation.}
Scene flow was first introduced by Vedula \etal~\cite{vedula1999three} as the dense 3D motion of all points in an observed scene from several calibrated cameras.
Several classical methods adopt energy minimization approaches, such as joint recovery of flow and stereo~\cite{Huguet:2007:Variational} and decoupled inference of stereo and flow for efficiency~\cite{Wedel:2008:Efficient}.
Compared with optical flow and stereo, the solution space of scene flow is of higher dimension and thus more challenging.
Vogel \etal~\cite{vogel2013piecewise} reduce the solution space by assuming a scene flow of piecewise rigid moving planes over superpixels.
Their work first tackles scene flow from a holistic perspective and outperforms contemporary stereo and optical flow methods by a large margin on the KITTI benchmark~\cite{Geiger:2012:We}.

\noindent\textbf{Joint scene understanding.}
Motion and segmentation are chicken-and-egg problems: knowing one simplifies the other. While the layered approach has long been regarded as an elegant solution to these two problems~\cite{Wang:1994:RMIL}, existing solutions tend to get stuck in local minima~\cite{Sun:NIPS:10}. In the {motion segmentation} literature, most methods start from an estimate of optical flow as input, and segment the scene by jointly estimating (either implicitly or explicitly) camera motion, object motion, and scene appearance, e.g.~\cite{bideau2018best, TokmakovCVPR17}. Lv \etal ~\cite{Lv18eccv} show that motion can be segmented directly from two images, without first calculating optical flow. Taylor \etal ~\cite{taylor2015causal} demonstrate that occlusion can also be a useful cue.

Exploiting advances in semantic segmentation,  Sevilla \etal~\cite{Sevilla-Lara16optical} show that semantic information is good enough to initialize the layered segmentation and thereby improves optical flow. Bai \etal~\cite{Bai2016Exploiting} use instance-level segmentation to deal with a small number of traffic participants. Hur and Roth~\cite{Hur:2016:Joint}  jointly estimate optical flow and temporally consistent semantic segmentation and obtain gains on both tasks. The object scene flow algorithm ~\cite{menze2015object} segments a scene into independently moving regions and enforces superpixels within each region to have similar 3D motion. The ``objects'' in their model are assumed to be planar and initialized via bottom-up motion estimation. %
Behl \etal~\cite{Behl2017ICCV}, Ren \etal~\cite{ren2017cascaded}, and Ma \etal~\cite{Ma2019DRISF} all show that instance segmentation helps scene flow estimation in the autonomous setting. While assuming a rigid motion for each individual instance works well for cars, this assumption tends to fail in general scenes, such as Sintel, on which our holistic approach achieves state-of-the-art performance.

The top-performing energy-based approaches are too computationally expensive for real-time applications.
Here we present a compact CNN model to holistically reason about geometry (disparity), motion (flow), and semantics, which runs much faster than energy-based approaches. %

\noindent\textbf{End-to-end learning of optical flow and disparity.} Recently %
CNN based methods have made significant progress on optical flow and disparity, two sub-problems of scene flow estimation.
Dosovitskiy \etal~\cite{Dosovitskiy:2015Flownet} first introduce two CNN models, FlowNetS and FlowNetC, for optical flow and bring about a paradigm shift to optical flow and disparity estimation. Ilg \etal~\cite{Ilg:2016:Flownet2} propose several technical improvements, such as dataset scheduling and stacking basic models into a big one, \ie, FlowNet2.
FlowNet2 has near real-time performance and obtains competitive results against hand-designed methods.
Ilg \etal~\cite{Ilg2018occlusions} stack networks for flow, disparity together for the joint task of scene flow estimation. However, there is no information sharing between the networks for flow and disparity.
Ranjan and Black \cite{Ranjan:2016:SpyNet} introduce a spatial pyramid network that performs on par with FlowNetC but has more than 100 times fewer parameters, due to the use of two classical principles: pyramids and warping.
Sun \etal~\cite{sun2018pwc} develop a compact yet effective network, called PWC-Net, which makes frequent use of three principles to construct the network: pyramids of learnable features, warping operations, and cost volume processing. PWC-Net obtains state-of-the-art performance on two major optical flow benchmarks.

The FlowNet work also inspired new CNN models for stereo estimation %
~\cite{Kendall_2017_ICCV,chang18pyramid,yang2018segstereo}. Kendall \etal~\cite{Kendall_2017_ICCV} concatenate features to construct the cost volume, followed by 3D convolutions. The 3D convolution becomes commonly-used for stereo but is computationally expensive in speed and memory. Chang and Chen~\cite{chang18pyramid}  introduce a pyramid pooling module to exploit context information for establishing correspondences in ambiguous regions. Yang \etal~\cite{yang2018segstereo} incorporate semantic cues to tackle textureless regions. Yin~\etal cast optical flow and disparity estimations as probabilistic distribution matching problems~\cite{yin2019hierarchical} to provide uncertainty estimation. They do not exploit the shared encoder of the two tasks as we do.

Existing scene flow networks~\cite{Ilg2018occlusions,Ma2019DRISF,mayer16a} stack independent networks for disparity and flow together. We are interested in exploiting the interactions among multiple related tasks to design a compact and effective network for holistic scene understanding.
Our holistic scene flow network  performs favorably against state of the art while being faster for inference and consuming less memory. In particular, we show the benefit of sharing the feature encoder between different tasks, such as flow and disparity.

\begin{figure*}[t]
\centering
\includegraphics[width=0.8\linewidth]{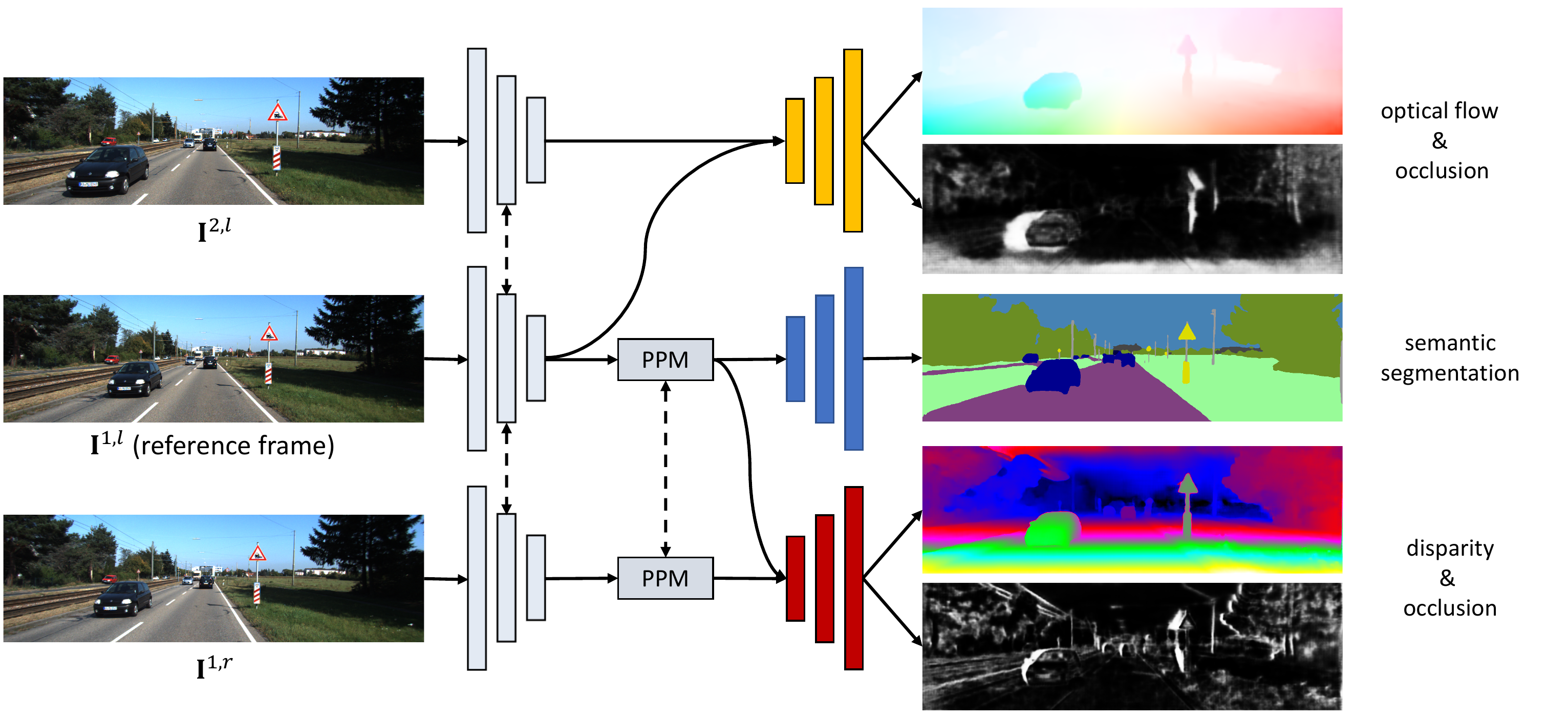}
\caption{Illustration of network design. Dashed arrows indicate shared weights. We have a single encoder for all input images and all different tasks and keep different decoders for different tasks. On the right, from top to bottom are: optical flow, forward occlusion mask, semantic segmentation, disparity, and disparity occlusion. The PPM (Pyramid Pooling Module) is not helpful for optical flow estimation. But thanks to the modular network design, we can flexibly configure the network.
}
\label{fig:modular_net_design}
\end{figure*}

\noindent\textbf{Self-supervised learning from videos.}
Supervised learning often uses synthetic data, as it is hard to obtain ground truth optical flow and disparity for real-world videos. Recently self-supervised learning methods have been proposed to learn scene flow by minimizing the data matching cost~\cite{Zou2018dfnet}
or interpolation errors~\cite{jiang2018super, liu17video}. However, the self-supervised methods have not yet achieved the performance of their supervised counterparts.

\section{Semi-Supervised Scene Flow Estimation}

We follow the problem setup of the KITTI scene flow benchmark~\cite{menze2015object}, as illustrated in Fig.~\ref{fig:modular_net_design}. The inputs are two stereo image pairs over time  $\left(\mv{I}^{1,l}, \mv{I}^{2,l}, \mv{I}^{1,r}, \mv{I}^{2,r}\right)$, %
where the first number in the superscript indicates the time step and the second symbol denotes the left or right camera. To save space, we will omit the superscript if the context is clear. %
We want to estimate optical flow $\mv{F}^{1,l}$ from the first left image to the second left image and disparity $\mv{D}^{1,l}$ and $\mv{D}^{2,l}$ from  the left image to the right image at the first and second frames, respectively. %
We also consider occlusion between two consecutive frames $\mv{O}_F^{1,l}$ and between the two sets of stereo images $\mv{O}_D^{1,l}$ and $\mv{O}_D^{2,l}$, as well as semantic segmentation for the reference (first left) image, \ie, $\mv{S}^{1,l}$. These extra outputs introduce interactions between different tasks to impose more constraints in the network training. Further, we hypothesize that sharing features among these closely-related tasks induces better feature representations.

We will first introduce our modular network design in Section~\ref{sec:model_design}, which shares an encoder among different tasks and supports flexible configurations during training.
We will then explain our semi-supervised loss function in Section~\ref{sec:partial_data}, which enables learning with partially labeled data.

\subsection{Modular Network Design}
\label{sec:model_design}

To enable feature sharing among different tasks and allow flexible configurations during training, we design the network in a modular way. Specifically, we build our network on top of PWC-Net~\cite{sun2018pwc}, a compact network for optical flow estimation. PWC-Net consists of an encoder and a decoder, where the encoder takes the input images and extracts features at different hierarchies of the network. The decoder is specially designed with domain knowledge of optical flow. The encoder-decoder structure allows us to design a network in a modular way, with a single shared encoder and several decoders for different tasks.

\noindent\textbf{Shared encoder.}
The original encoder of PWC-Net, however, is not well-suited to multiple tasks because of its small capacity. More than 80\% of the parameters of PWC-Net are concentrated in the decoder, which uses DenseNet~\cite{huang17dense} blocks at each pyramid level. The encoder consists of plain convolutional layers and uses fewer than 20\% of the parameters. While sufficient for optical flow, the encoder does not work well enough for disparity estimation. To make the encoder versatile for different tasks, we make the following modifications.
First, we reduce the number of feature pyramid levels from 6 to 5, which reduces the number of parameters by nearly 50\%. It also allows us to borrow the widely-used 5-level ResNet-like encoder architecture~\cite{chang18pyramid, he16deep}, which has been proven to be effective in a variety of vision tasks. Specifically, we replace plain CNN layers with residual blocks~\cite{he16deep} and add Batch Normalization layers~\cite{ioffe15batch} in both encoder and decoder. %
With these modifications, the new model has slightly fewer parameters but gives better disparity estimation results and also better flow (Table~\ref{tab:flow_sintel}).

\noindent\textbf{Decoder for disparity.} Next we explain how to adapt PWC-Net to disparity estimation between two stereo images.
Disparity is a {special case} of optical flow computation, with correspondences lying on a horizontal line. As a result, we need only to build a 1D cost volume for disparity, while the decoder of the original PWC-Net constructs a 2D cost volume for optical flow. Specifically, for optical flow, a feature at $p\!=\!(x, y)$ in the first feature map is compared to features at $q\in[x\!-\!k, x\!+\!k]\!\times\![y\!-\!k, y\!+\!k]$ in the warped second feature map. For disparity, we need only to search for correspondences by comparing $p$ in the left feature map to $q\in[x\!-\!k, x\!+\!k]\!\times\! y$ in the warped right feature map. We use $k\!=\!4$ for both optical flow and disparity estimations. %
Across the feature pyramids, our decoder for disparity adopts the same warping and refinement process as PWC-Net.

To further improve disparity estimation accuracy, we investigate more design choices. First, we use the Pyramid Pooling Module (PPM)~\cite{zhao17pyramid} to aggregate the learned features of input images across multiple levels. Second, the decoder outputs a disparity map one fourth the size of the input resolution, which tends to have blurred disparity boundaries. As a remedy, we add a simple hourglass module widely used in disparity estimation~\cite{chang18pyramid}. It takes a twice upsampled disparity, a feature map of the first image, and a {warped} feature map of the second image to predict  a residual disparity that is added to the upsampled disparity. Both the PPM and hourglass modifications lead to significant improvements in disparity estimation.
They are not helpful for optical flow estimation though, indicating that the original PWC-Net is well designed for optical flow.
The modular design allows us to flexibly configure networks that work for different tasks, as shown in Fig.~\ref{fig:modular_net_design}.

\noindent\textbf{Decoder for segmentation.}
To introduce more constraints to network training, we also consider semantic segmentation. It encourages the encoder to learn some semantic information, which may help optical flow and disparity estimations. For semantic segmentation decoder, we use the UPerNet~\cite{xiao18unified} for its simplicity.

\noindent\textbf{Occlusion estimation}. For occlusion predictions, we add sibling branches to optical flow or disparity decoders to perform pixel-wise binary classification, where $1$ means fully occluded. Adding such extra modules enables holistic scene understanding that helps us to induce better feature representations in the shared encoder and use extra supervision signals for network training to deal with partially labeled data, which is discussed in Section~\ref{sec:partial_data}. Critically, for scene flow estimation, the shared encoder results in a more {compact} and efficient model. For optical flow and disparity estimations, we can combine modules as needed during training, with no influence on the {inference time}. For scene flow estimation, extra modules can be used optionally, depending on configuration. See explanations in Section~\ref{sec:main_results}.

\begin{figure}
\centering
\renewcommand{\tabcolsep}{0.25mm}
\begin{tabular}{ccc}
\includegraphics[width=0.33\linewidth]{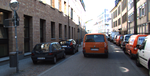} &
\includegraphics[width=0.33\linewidth]{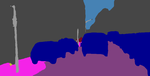} &
\includegraphics[width=0.33\linewidth]{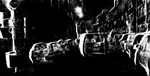} \\
{\small Left input image} & {\small Pre-trained seg.} & {\small Pre-trained occ.} \\
\includegraphics[width=0.33\linewidth]{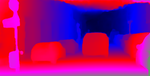} &
\includegraphics[width=0.33\linewidth]{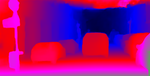} &
\includegraphics[width=0.33\linewidth]{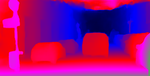} \\
{\small Supervised loss} & {\small + segmentation loss} & {\small + occlusion loss} \\
\includegraphics[width=0.33\linewidth]{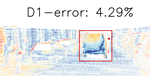} &
\includegraphics[width=0.33\linewidth]{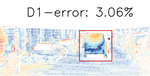} &
\includegraphics[width=0.33\linewidth]{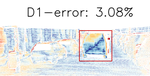} \\
\multicolumn{3}{c}{{\small Disparity error map (blue lower error, red higher error)}}  \\
\includegraphics[width=0.33\linewidth]{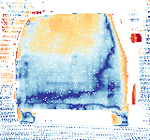} &
\includegraphics[width=0.33\linewidth]{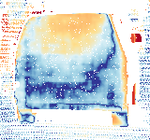} &
\includegraphics[width=0.33\linewidth]{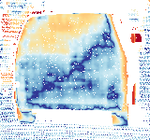} \\
\multicolumn{3}{c}{{\small Enlarged view of error map for the car (best viewed in color)}}  \\
\end{tabular}
\caption{Effects of adding distillation losses for semantic segmentation (middle) and occlusion (right) to the supervised loss.
}
    \label{fig:occ_distill_loss}
\end{figure}

\subsection{Semi-Supervised Loss}
\label{sec:partial_data}

No fully labeled datasets are available to directly train our holistic scene flow network. For example, KITTI has no ground-truth occlusion masks. Even for optical flow and disparity ground-truths, only around 19\% of pixels of the KITTI data have annotations due to the difficulty in data capturing. The synthetic SceneFlow dataset~\cite{mayer16a} has no ground truth for semantic segmentation. To address these issues, we introduce our semi-supervised loss functions, which consist of supervised, distillation, and self-supervised loss terms.

\newcommand\px{\mathbf{x}}
\newcommand{\calL}{\mathcal{L}}

\noindent\textbf{Supervised loss.} When corresponding ground-truth annotations are available, we define our {supervised} loss as
\aroundEqn
\begin{align}
    \calL_{sp} = \left(\calL_F + \calL_{O_F}\right) +  \left(\calL_D + \calL_{O_D}\right),
\end{align}
\aroundEqn
where $\calL_F$ and $\calL_{O_F}$ are loss terms for estimating optical flow and its corresponding occlusion. $\calL_D$ and $\calL_{O_D}$ are the loss terms for estimating disparity and its corresponding occlusion. $\calL_F$ is defined across multiple pyramid levels as
\aroundEqn
\begin{align}
    \calL_F = &\sum_{i=1}^{N_F} \omega_i \sum_p \rho\left( \mv{F}_i(p), \hat{\mv{F}}_i(p)\right),
\end{align}
\aroundEqn
where $\omega_i$ denotes optical flow and disparity weights at pyramid level $i$, $N_F$ is the number of pyramid levels, and  $\rho(\cdot, \cdot)$ is a loss function measuring the similarity between the ground-truth $\mv{F}_i(p)$ and estimated optical flow $\hat{\mv{F}}_i(p)$ at pixel $p$. Disparity and occlusion loss functions, $\calL_D$, $\calL_{O_F}$, and $\calL_{O_D}$ are defined in a similar way. We use $L_2$ and {\texttt{smooth\_l1}}~\cite{girshick15fast,chang18pyramid} loss for optical flow and disparity estimations, respectively.
For the occlusions, we use binary cross entropy loss when ground-truth annotations are available (\eg, on FlyingThings3D~\cite{Mayer:2016:Large}).
For semantic segmentation, only ground-truth annotations of the left images are available for KITTI2015. We empirically found using distillation loss only introduced below yields better accuracy.

\begin{figure}
\centering
\renewcommand{\tabcolsep}{0.25mm}
\begin{tabular}{ccc}
\includegraphics[width=0.33\linewidth]{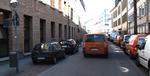} &
\includegraphics[width=0.33\linewidth]{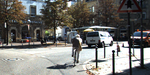} &
\includegraphics[width=0.33\linewidth]{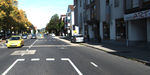} \\
\includegraphics[width=0.33\linewidth]{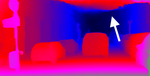} &
\includegraphics[width=0.33\linewidth]{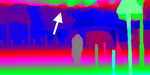} &
\includegraphics[width=0.33\linewidth]{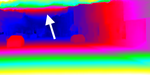} \\
\includegraphics[width=0.33\linewidth]{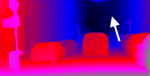} &
\includegraphics[width=0.33\linewidth]{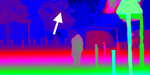} &
\includegraphics[width=0.33\linewidth]{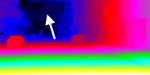} \\
\end{tabular}
\caption{Illustration of effectiveness of self-supervised loss. From top to bottom: input images, disparity estimations \emph{without} using self-supervised loss, and disparity estimations \emph{with} using self-supervised loss. We can see self-supervised loss helps greatly reduce artifacts in the sky region.}
\label{fig:effect_ss_loss}
\end{figure}

\noindent\textbf{Distillation loss.} For occlusion estimation and semantic segmentation tasks, ground-truth annotations are not always available. They are important, however, during network training. For instance, on KITTI, supervised loss can only be computed on sparsely annotated pixels. Adding extra supervision for occlusion estimation is helpful for the network to extrapolate optical flow and disparity estimations to regions where ground-truth annotations are missing, yielding visually appealing results.

We find the occlusion estimations provided by a pre-trained model on synthetic data are reasonably good, as shown in Fig.~\ref{fig:occ_distill_loss}. As a {soft supervision}, we encourage the occlusion estimations of the network during training do not deviate much from what it learned in the pre-training stage. Therefore, we simply use the estimations of a pre-trained network as pseudo ground-truth and {\texttt{smooth\_l1}} loss function during training, computed in multiple pyramid levels as $\calL_F$ and $\calL_D$. Adding extra supervision using distillation loss for occlusion is helpful for reducing artifacts in disparity estimation, as shown in Fig.~\ref{fig:occ_distill_loss}.

For semantic segmentation, we use the distillation loss formulation proposed in~\cite{hinton15distilling}. Specifically, semantic segmentation distillation loss $\calL_{S_d}$ for a single pixel $p$ (omitted here for simplicity) is defined as
\aroundEqn
\begin{align}
    \calL_{S_d} = T\sum_{i=1}^C \tilde{y}_i\log \hat{y}_i,~~~\tilde{y}_i=\frac{\exp^{-z_i/T}}{\sum_{k}\exp^{-z_{k}/T}},
\label{eq:seg_distill_loss}
\end{align}
\aroundEqn
where $C$ is the number of segmentation categories. $z_i$ and $\tilde{y}_i$ come from a more powerful {teacher} segmentation model, where $z_i$ is the output for the $i$-th category right before the \texttt{softmax} layer, also known as \texttt{logit}. $\tilde{y}_i$ is ``softened'' posterior probability for the $i$-th category, controlled by the hyper-parameter $T$~\cite{hinton15distilling}. We empirically found $T\!=\!1$ works well on a validation set.  $\hat{y}_i$ is the estimated posterior probability of our model. The distillation is aggregated over all pixels in training images.

\noindent\textbf{Self-supervised loss.} To further constrain the network training, we also define self-supervised loss. Optical flow and disparity are defined as correspondence between two input images. We can therefore compare two corresponding pixels defined by either optical flow or disparity as supervision for network training.

The most straightforward metric is to compare values between two corresponding pixels that are visible in both frames, known as {photometric consistency}. In a single pyramid level, it is defined as $\calL_{PC}\!=\!$ %
\aroundEqn
\begin{align}
\|\mv{I}^l \!-\! g(\mv{I}^r, \mv{D}^l) \|_1 \!\odot\! \bar{\mv{O}}_D \!+\!  \| \mv{I}^1\!-\!g(\mv{I}^2, \mv{F}^1) \|_1 \!\odot\! \bar{\mv{O}}_F, %
\label{eqn:pht_cons}
\end{align} \aroundEqn
where $g(\cdot, \cdot)$ is the differentiable warping function,  $\bar{\mv{O}}\!=\!1\!-\!\mv{O}$, $\odot$ denotes element-wise multiplication followed by summation, and we omit some superscripts when the context is clear. This loss term reasons about occlusion by modulating the consistency loss using the occlusion map and tightly couples occlusion with optical flow and stereo.

\newcommand{\py}{\mathbf{y}}
As photometric consistency is not robust to lighting changes, we further introduce {semantic consistency}, encouraging two corresponding pixels to have similar semantic segmentation posterior probability. Specifically, this semantic consistency is defined as $\calL_{SC} \!=\! $ %
\aroundEqn
\begin{align}
\|\tilde{\py}^l \!-\! g(\tilde{\py}^r,\!\mv{D}^l) \|_1 \!\odot\! \bar{\mv{O}}_D \!+\! \|\tilde{\py}^1\!-\!g(\tilde{\py}^2,\!\mv{F}^1)\|_1 \!\odot\! \bar{\mv{O}}_F, \!\!\!\!
\label{eqn:sem_cons}
\end{align}\aroundEqn
where $\tilde{\py}$ denotes a posterior probability image coming from the teacher segmentation network used in Eq.(\ref{eq:seg_distill_loss}). Unlike raw pixel values, the segmentation posterior probability is more robust to lighting changes.

Finally, we consider the structural similarity loss %
\aroundEqn
\begin{align}
\calL_{SS} \!=\!& \gamma_D \left(1 \!-\! SS(\mv{I}^l, \mv{I}^l\otimes \mv{O}_D + g(\mv{I}^r, \mv{D}^l)\otimes \bar{\mv{O}}_D) \right)  + \nonumber \\
& \gamma_F \left(1\!-\!SS(\mv{I}^1,\mv{I}^1\otimes\mv{O}_F + g(\mv{I}^2, \mv{F}^1)\otimes \bar{\mv{O}}_F)\right),
\end{align}
\aroundEqn
where $\otimes$ indicates element-wise multiplications only. $SS(\cdot, \cdot)$ is a differentiable function that outputs a single scalar value to measure the structural similarity between two input images~\cite{zhao17loss}. Note that for occluded pixels in the warped image, their values are replaced with values of pixels at the same position in the left/first image.

There exist trivial solutions for minimizing Eq.(\ref{eqn:pht_cons}) and Eq.(\ref{eqn:sem_cons}) by setting $\mv{O}_D$ and $\mv{O}_F$ to all ones. We thus add regularization terms
\aroundEqn
\begin{align}
\calL_{REG}=\beta_D\sum_p \mv{O}_D(p) + \beta_F\sum_p \mv{O}_F(p),
\end{align}
\aroundEqn

Although the self-supervised photometric and structural similarity loss terms have been studied in previous work~\cite{jason2016back2basics,godard17unsupervised}, our definition differs from theirs in that we model occlusions. On one hand, we avoid defining loss terms in the occluded regions. On the other hand, these self-supervised terms provide modulation for the occlusion estimation as well. Thus, our networks tightly couple these four closely-related tasks together.

Our final semi-supervised loss consists of supervised, distillation, and self-supervised loss terms. More details can be found in the supplementary material.

\begin{table}[t]
\caption{Average EPE results on MPI Sintel optical flow dataset.  ``-ft'' means fine-tuning on the MPI Sintel \emph{training} set and the numbers in parentheses are results on the data the methods have been fine-tuned on.
}
\vspace{-1mm}
\label{tab:flow_sintel}
\footnotesize
\centering
\begin{tabular}{lcccccc} \\
    \toprule
	\multirow{2}*{Methods} & \multicolumn{2}{c}{Training}  & \multicolumn{2}{c}{Test} & Time\\
	& Clean &  Final & Clean &  Final &  (s) \\ \hline
	FlowFields~\cite{Bailer2015Flow} & -& -& 3.75  & 5.81  & 28.0 \\
	MRFlow~\cite{Wulff2017Optical} & 1.83 & 3.59  &  \bd{2.53} & 5.38 & 480\\
	FlowFieldsCNN ~\cite{Bailer_2017_CVPR} & - & - & 3.78 & 5.36 &23.0\\
	DCFlow~\cite{Xu2017Accurate}& - & - & 3.54 & \ud{5.12} & 8.60\\
	\midrule
	SpyNet-ft~\cite{Ranjan:2016:SpyNet} & (3.17) & (4.32) & 6.64 & 8.36  &0.16\\
	FlowNet2~\cite{Ilg:2016:Flownet2} & 2.02 &   3.14	 &   3.96 & 6.02 &0.12\\
	FlowNet2-ft~\cite{Ilg:2016:Flownet2} & ({1.45})	&  ({2.01}) &  4.16 & 5.74 &0.12	\\ %
    LiteFlowNet~\cite{Hui_2018_CVPR}  & (1.64) & (2.23) & 4.86 & 6.09 &0.09 \\
	PWC-Net~\cite{sun2018pwc} & 2.55 &	3.93 & - & - & \bd{0.03}\\
	PWC-Net-ft~\cite{sun2018pwc} & ({1.70})	& ({2.21}) & {3.86}  & 5.13    & \bd{0.03}  \\
	FlowNet3~\cite{Ilg2018occlusions} &  2.08 & 3.94 & 3.61 & 6.03 & 0.07 \\
	FlowNet3-ft~\cite{Ilg2018occlusions} &  (\bd{1.47}) & (2.12) & 4.35 & 5.67 & 0.07 \\
	\midrule
	SENSE & 1.91 & 3.78 & - & - & \bd{0.03} \\
	SENSE-ft & (1.54) & (\bd{2.05}) & 3.60 & \bd{4.86} & \bd{0.03} \\
	\bottomrule
\end{tabular}
\end{table}

\section{Experiments}

\subsection{Implementation Details}
\noindent \textbf{Pre-training of stereo and optical flow.}
We use the synthetic SceneFlow dataset~\cite{Mayer:2016:Large}, including FlyingThings3D, Monkaa, and Driving, for pre-training. All three datasets contain optical flow and disparity ground-truth. Occlusion labels are only available in FlyingThings3D. During training, we uniformly sample images from all three datasets and compute occlusion loss when the ground-truths are available.
During training, we use color jittering for both optical flow and disparity training. Additionally, we use random crops and vertical flips for stereo training images. The crop size is $256\times 512$. For optical flow training images, we perform extensive data augmentations  including random crop, translation, rotation, zooming, squeezing, and horizontal and vertical flip, where the crop size is $384\times 640$.
The network is trained for 100 epochs with a batch size of 8 using the Adam optimizer~\cite{kingma14adam}. We use synchronized Batch Normalization~\cite{xiao18unified} to ensure there are enough training samples for estimating Batch Normalization layers' statistics when using multiple GPUs. The initial learning rate is 0.001 and decreased by factor of 10 after 70 epochs.

\begin{table}[t]
\caption{Results on the KITTI optical flow dataset. ``-ft'' means fine-tuning on the KITTI \emph{training} set and the numbers in the parenthesis are results on the data the methods have been fine-tuned on.} 	\vspace{-1mm}
\label{tab:flow_kitti}
\footnotesize
\centering
\setlength\tabcolsep{2pt}
\begin{tabular}{lcccccc} \\
    \toprule
	\multirow{3}*{Methods} & \multicolumn{3}{c}{ KITTI 2012} & \multicolumn{3}{c}{ KITTI 2015}  \\
	& AEPE & AEPE & Fl-Noc &  AEPE   & Fl-all & Fl-all \\
	& \emph{train} & \emph{test} & \emph{test}   & \emph{train} & \emph{train}  & \emph{test} \\  \hline
	FlowFields~\cite{Bailer2015Flow} & -& -& -& - &-  & 19.80\%  \\
	MRFlow~\cite{Wulff2017Optical} & - &  -&- &-  &  14.09 \% & 12.19 \% \\ %
	DCFlow~\cite{Xu2017Accurate} & - &  -&- &- & 15.09 \% &14.83 \% \\ %
	SDF~\cite{Bai2016Exploiting} & - &2.3 & {3.80}\% & - &- & 11.01 \% \\ %
	MirrorFlow~\cite{Hur_2017_ICCV} &- & 2.6 & 4.38\% & - & 9.93\% & \ud{10.29}\% \\ %
	\midrule
	SpyNet-ft~\cite{Ranjan:2016:SpyNet} & (4.13)  & 4.7 &12.31\% & - & - & 35.07\%  \\ %
	FlowNet2~\cite{Ilg:2016:Flownet2}  &4.09& - & - & 10.06 & 30.37\% & - \\ %
	FlowNet2-ft~\cite{Ilg:2016:Flownet2}  & ({1.28}) & {1.8} & 4.82\% & (\ud{2.30}) & ({8.61}\%) & {10.41} \% \\ %
	LiteFlowNet~\cite{Hui_2018_CVPR}  & ({1.26}) & {1.7} & - & ({2.16}) & ({8.16}\%) & {10.24} \% \\
	PWC-Net~\cite{sun2018pwc} & 4.14 & - &- & 10.35 & 33.67\% &-\\ %
	PWC-Net-ft~\cite{sun2018pwc} & ({1.45})	&  {1.7}	 &4.22\%	& ({2.16})	& ({9.80}\%) & {9.60}\%  		\\
	FlowNet3~\cite{Ilg2018occlusions} & 3:69 & - & - & 9.33 & - & - \\
    FlowNet3-ft~\cite{Ilg2018occlusions} & (1.19) & - & 3.45\% & (\bd{1.79}) & - & 8.60\% \\
	\midrule
	SENSE & 2.55 & - & - & 6.23 & 23.29\% & - \\
	SENSE-ft & (\bd{1.14}) & \bd{1.5} & \bd{3.00}\% & (2.01) & (\bd{9.20}\%) & 8.38\% \\
	SENSE-ft+semi & (1.18) & 1.5 & 3.03\% & (2.05) & (9.69\%) & \bd{8.16}\% \\
	\bottomrule
\end{tabular}
\end{table}

\noindent \textbf{Fine-tuning}. For Sintel, we use a similar learning rate schedule as used in~\cite{sun2018pwc}. On KITTI 2012~\cite{Geiger:2012:We} and KITTI 2015~\cite{Menze2015ISA}
, we use longer learning rate schedule, where the model is trained for 1.5K epochs with an initial learning rate is 0.001. We perform another 1K-epoch training with an initial learning rate of 0.0002.
We use a crop size of $320\times 768$ for both disparity and optical flow training images and a batch size of 8. More training details are provided in the supplementary material due to limited space here.

\noindent \textbf{Training semantic segmentation}. We jointly train all parts of the entire network, including pre-trained encoder and decoders for optical flow and disparity, as well as a randomly initialized segmentation decoder. We empirically found using a randomly initialized segmentation decoder yields better performance.

\begin{table}
\caption{Results on KITTI stereo datasets (test set).
 }
\label{tab:disp_kitti}
\setlength\tabcolsep{1.4pt}
\footnotesize
\centering
\begin{tabular}{l|cc|cccc|c}
\toprule
\multirow{3}{*}{Methods}
& \multicolumn{2}{c|}{KITTI 2012} & \multicolumn{4}{c|}{KITTI 2015} & \multirow{3}{*}{Time} \\
 & All & Non-Occ & \multicolumn{2}{c}{All}& \multicolumn{2}{c|}{Non-Occ} & \\
 \cmidrule(lr){2-2}\cmidrule(lr){3-3}\cmidrule(lr){4-5}\cmidrule(lr){6-7}
 & Out-All & Out-Noc & D1-fg & D1-all & D1-fg & D1-all &  (s) \\
\midrule
\footnotesize{Content-CNN}~\cite{luo2016efficient} & 3.07 & 4.29 & 8.58  & 4.54 & 7.44  & 4.00 & 1.0 \\
DispNetC \cite{Mayer:2016:Large} & - & - & {4.41}  & 4.34 & {3.72}  & 4.05 & \bd{0.06}\\
MC-CNN \cite{zbontar2016stereo} & 2.43 & 3.63 & 8.88  & 3.89 & 7.64  & 3.33 & 67  \\
PBCP \cite{Seki2016BMVC} & 2.36 & 3.45 & 8.74  & 3.61 & 7.71  & 3.17 & 68  \\
Displets v2 \cite{guney2015displets} & 2.37 & 3.09 & 5.56  & 3.43 & 4.95  & 3.09 & 265 \\
\midrule
GC-Net~\cite{Kendall_2017_ICCV}	& 1.77 & 2.30 & 6.16  & {2.87} & 5.58  & {2.61} & 0.9 \\
PSMNet~\cite{chang18pyramid} & \bd{1.49} & \bd{1.89} & 4.62 & 2.32 & 4.31 & 2.14 & 0.41 \\
SegStereo~\cite{yang2018segstereo} & 1.68 & 2.03 & 3.70 & \bd{2.08} & 4.07 & 2.25 & 0.6 \\
FlowNet3~\cite{Ilg2018occlusions} & 1.82 & - & - & 2.19 & - & - & 0.07 \\
\midrule
SENSE & 1.77 & 2.18 & 3.13 & 2.33 & 2.79 & 2.13 & \bd{0.06} \\
SENSE+semi & 1.73 & 2.16 & \bd{3.01} & 2.22 & \bd{2.76} & \bd{2.05} & \bd{0.06}\\
\bottomrule
\end{tabular}
\end{table}

For the segmentation distillation loss and semantic consistency loss computation, we first train the teacher segmentation model. We use the ResNet101-UPerNet~\cite{xiao18unified} pre-trained on CityScapes~\cite{Cordts2016Cityscapes} using its training set with fine annotations only, which achieves 75.4\% IoU on the validation set. We fine-tune the model on KITTI 2015~\cite{Alhaija2018IJCV}, where the segmentation annotations, consistent with CityScapes' annotation style, for the left images are provided.

\subsection{Main Results}
\label{sec:main_results}

\noindent\textbf{Optical flow results.} Table~\ref{tab:flow_sintel} shows the results for optical flow estimation on the MPI Sintel benchmark dataset. Our approach outperforms CNN-based approaches without or with fine-tuning. On the more photorealistic (final) pass of the test set, which involves more rendering details such as lighting change, shadow, motion blur, etc, our approach outperforms both CNN-based and traditional hand-designed approaches by a large margin.

Table~\ref{tab:flow_kitti} shows the results on both KITTI2012 and KITTI2015. Our approach significantly outperforms both hand-designed and CNN-based approaches on KITTI 2012 with and without fine-tuning. On KITTI 2015, our model achieves much lower error rates than CNN-based approaches without pre-training (including ours).
After fine-tuning, it outperforms all other approaches. %

We note that better optical flow results are reported in an improved version of PWC-Net~\cite{sun2019models}, which uses FlyingChairs followed by FlyingThings3D for pre-training. It also uses much longer learning rate schedules for fine-tuning, so the results are not directly comparable to ours.

\begin{table}[t]
\centering
\footnotesize
\caption{Results on KITTI2015 Scene flow dataset. CNN-based approaches need to deal with refinement of D2, where N and R indicates network and rigidity-based refinement, respectively.}
\label{tab:scene_kitti2015}
\renewcommand{\tabcolsep}{1.5mm}
\begin{tabular}{l| c | c | c | c| c|c}
\toprule
Method &  D1-all &  D2-all & Fl-all & SF-all & D2 ref. & Time (s) \\
\midrule
ISF  \cite{Behl2017ICCV}                         &   $4.46$      &   $5.95$      &   $6.22$        &  $8.08$    & -    & $600$ \\
\midrule
CSF \cite{Lv2016ECCV} & 5.98 & 10.06 & 12.96 & 15.71 & -  & 80 \\
SGM+FF\cite{schuster2018combining}      &   $13.37$     &   $27.80$     &   $22.82$    &  $33.57$   & -  & $29$ \\
SceneFF\cite{schuster2018sceneflowfields}      &   $6.57$      &   $10.69$     &   $12.88$    &  $15.78$   & -  & $65$\\
FlowNet3~\cite{Ilg2018occlusions} &   $2.16$      &   $6.45$      &   $8.60$     &  $11.34$  & N   & $0.25$ \\
\midrule
SENSE & 2.23 & 7.37 & 8.38 & 11.71 & N & 0.16 \\
SENSE+semi & 2.22 & 6.57 & 8.16 & 11.35 & N & 0.16 \\
SENSE+semi & 2.22 & 5.89 & 7.64 & 9.55 & R+N & 0.32 \\
\bottomrule
\end{tabular}
\end{table}

\noindent\textbf{Disparity results.} For disparity estimation, SENSE significantly outperforms previous CNN-based approaches including DispNetC \cite{Mayer:2016:Large} and GC-Net \cite{Kendall_2017_ICCV} and achieves comparable accuracy with state-of-the-art approaches like PSMNet~\cite{chang18pyramid}, SegStereo~\cite{yang2018segstereo}, and FlowNet3~\cite{Ilg2018occlusions}. Notably, our approach performs the best on the foreground region in both all and non-occluded regions on KITTI2015.

\noindent\textbf{Scene flow results.} Table~\ref{tab:scene_kitti2015} shows Scene flow results on KITTI 2015. SENSE performs the best in general CNN-based scene flow methods, compared to FlowNet3 \cite{Ilg2018occlusions}. Compared to ISF~\cite{Behl2017ICCV}, SENSE is 2K times faster and can handle general nonrigid scene motions.

To remove artifacts introduced by the second frame disparity warping operation, we use a refinement network of a encoder-decoder structure with skip connections. It takes $\mv{I}^{1,l}$, $\mv{O}_F^{1,l}$, $\mv{D}^{1,l}$, and $g(\mv{D}^{2,l}, \mv{F}^{1,l})$ to generate a residual that is added to the warped disparity. From our holistic outputs, we can refine the background scene flow using a rigidity refinement step. We first determine the static rigid areas according to semantic segmentation outputs. We then calculate the ego-motion flow by minimizing the geometry consistency between optical flow and disparity images using the Gauss-Newton algorithm. Finally, we compute the warped scene flow using the disparity of the reference frame and the ego-motion to substitute the raw scene flow only in the rigid background region.
This step additionally produces camera motion and better scene flow with minimal costs. Details of refinement steps are provided in supplementary material.

\noindent\textbf{Running time.} SENSE is an efficient model. SENSE takes 0.03s to compute optical flow between two images of size 436$\times$1024. For disparity, SENSE is an order of magnitude faster than PSMNet and SegStereo, and slightly faster than FlowNet3. For scene flow using KITTI images, SENSE takes 0.15s to generate one optical flow and two disparity maps. The additional warping refinement network takes 0.01s and the rigidity refinement takes 0.15s.

\noindent \textbf{Model size and memory.} SENSE is small in size. It has only 8.8M parameters for the optical flow model, and 8.3M for the disparity model. The scene flow model with shared encoder has 13.4M parameters. In contrast, FlowNet3 has a flow model (117M) and a disparity model (117M), which is 20 times larger. SENSE also has a low GPU memory footprint. FlowNet3 costs 7.4GB while SENSE needs 1.5GB RAM only. Although PSMNet has fewer parameters (5.1M), it costs 4.2GB memory due to 3D convolutions.

\begin{table}[t]
    \centering
    \footnotesize
    \caption{Effectiveness of different tasks.
    }
    \label{tab:effect_tasks}
    \setlength\tabcolsep{2pt}
    \begin{tabular}{ccc|ccc}
    \toprule
    \multicolumn{3}{c|}{Tasks} & \multicolumn{3}{c}{Results} \\
    \cmidrule(lr){1-3}\cmidrule(lr){4-6}
    flow & disp & seg & flow (F1-occ) $\downarrow$ & disp (D1-occ) $\downarrow$ & seg (mIoU) $\uparrow$ \\
    \midrule
    \checkmark & & & 11.37\% & - & - \\
    & \checkmark &  & - & 2.73\% & - \\
    & & \checkmark & - & - & 47.51\% \\
    \midrule
    \checkmark & \checkmark & & 11.59\% & 2.61\% & - \\
    \checkmark & & \checkmark & 11.39\% & - & 49.54\% \\
     & \checkmark & \checkmark & - & 2.62\% & 49.12\% \\
    \midrule
    \checkmark & \checkmark & \checkmark & 11.19\% & 2.59\% & 48.25\% \\
    \bottomrule
    \end{tabular}
\vspace{-12pt}
\end{table}

\subsection{Ablation Studies}

\newcommand{\subfigimg}[3][,]{%
	\setkeys{Gin,subfigpos}{vsep,hsep,#1}%
	\setbox1=\hbox{\includegraphics{#3}}%
	\leavevmode\rlap{\usebox1}%
	\rlap{\hspace*{8pt}\raisebox{\dimexpr\ht1-9pt}{\figlabel{#2}}}%
	\phantom{\usebox1}%
}

\begin{table}[t]
\caption{Ablation study of different loss terms. } %
\label{tab:ablations_loss}
\footnotesize
\renewcommand{\tabcolsep}{5pt}
\centering
\begin{tabular}{ccccc|ccccc}
\toprule
\multicolumn{2}{c}{Distillation } & \multicolumn{3}{c|}{Self-supervised} & \multicolumn{1}{c}{Flow} & \multicolumn{1}{c}{Disp} & Seg \\
\cmidrule(lr){1-2}\cmidrule(lr){3-5}\cmidrule(lr){6-6}\cmidrule(lr){7-7}\cmidrule(lr){8-8}
seg. & occ. & sem. & pho. & ss  & F1-Occ$\downarrow$  & D1-Occ$\downarrow$ & mIoU$\uparrow$\\
\midrule
& & & & &  11.16\%  & 2.52\% & - \\
\midrule
\checkmark & & & & & 10.96\% & 2.44\% & 51.48\% \\
& \checkmark & & & & 11.07\% & 2.38\% & - \\
\checkmark & \checkmark & & & & 11.17\% & 2.33\% & 51.26\% \\
\midrule
& & \checkmark & & & 11.11\% & 2.38\% & -\\
& & & \checkmark & & 11.04\% & 2.55\% & - \\
& & & & \checkmark & 11.16\% & 2.47\% & - \\
& &  \checkmark & \checkmark & \checkmark & 11.21\% & 2.58\% & - \\
\midrule
\checkmark & \checkmark & \checkmark & \checkmark & \checkmark & 11.12\% & 2.49\% & 50.92\% \\
\bottomrule
\end{tabular}
\vspace{-18pt}
\end{table}

\noindent\textbf{Performance of different tasks.} We report results of different tasks using different combinations of encoder and decoders. Our models are trained using 160 images of KITTI 2015 with a half of the aforementioned learning rate schedule. Results are reported on the rest 40 images in Table~\ref{tab:effect_tasks}. We can see that the shared encoder model performs better than models trained separately.

\noindent\textbf{Semi-supervised loss.} To study the effects of distillation and self-supervised loss terms, we perform ablation studies using all images of KITTI 2012 and 160 images of KITTI 2015 for training with a half of full learning rate schedule. The rest 40 ones of KITTI 2015 are used for testing. We finetune the baseline model using sparse flow and disparity annotations only. Table~\ref{tab:ablations_loss} shows the quantitative comparisons and Fig.~\ref{fig:effect_ss_loss} highlights the effects qualitatively.

Regarding distillation loss, both segmentation and occlusion distillation loss terms are useful for disparity and optical flow estimation. However, distillation loss is not helpful for reducing the artifacts in sky regions. Thus, the self-supervised loss is essential, as shown in Fig.~\ref{fig:effect_ss_loss}, though  quantitatively self-supervised loss is not as effective as the distillation loss. Finally, combining all loss terms yields the best optical flow and disparity accuracies.
We also test SENSE trained using semi-supervised loss on KITTI, as summarized in Tables~\ref{tab:flow_kitti}, \ref{tab:disp_kitti}, and \ref{tab:scene_kitti2015}. We can see it
improves disparity and optical flow accuracy on KITTI 2015 and also leads to better disparity on KITTI 2012.

\section{Conclusion}
\vspace{-6pt}
We have presented a compact network for four closely-related tasks in holistic scene understanding:
Sharing an encoder among these  tasks not only makes the network compact but also improves performance by exploiting the interactions among these tasks. It also allows us to introduce distillation and self-supervision losses to deal with partially labeled data.
Our holistic network has similar accuracy and running time as specialized networks for optical flow. It performs favorably against state-of-the-art disparity and scene flow methods while being much faster and memory efficient. Our work shows the benefits of synergizing
closely-related tasks for holistic scene understanding and we hope the insights will aid new research in this direction.

{\footnotesize
\section*{Acknowledgement}
\vspace{-6pt}
Huaizu Jiang and Erik Learned-Miller acknowledge support from AFRL and DARPA (\#FA8750-
18-2-0126) and the MassTech Collaborative grant for funding the UMass GPU cluster.
The U.S. Gov. is authorized to reproduce and distribute reprints for Gov.
purposes notwithstanding any copyright notation thereon. The views and
conclusions contained herein are those of the authors and should not be interpreted as necessarily representing the official policies or endorsements,
either expressed or implied, of the AFRL and DARPA or the U.S. Gov.
}

{\small
\bibliographystyle{ieee_fullname}
\bibliography{egbib}
}

\appendixtitleon
\appendixtitletocon
\begin{appendices}

\onecolumn

\section{Training Details}
We perform pre-training on the synthetic SceneFlow dataset and fine-tuning on Sintel and KITTI, respectively.

\noindent\textbf{Synthetic SceneFlow Dataset.} We use the subset of FlyingThings3D used in~\cite{Ilg2018occlusions}, Monkaa, and Driving for pre-training. We remove images whose maximum optical flow magnitude is greater than 500. We end up using 128,753 samples for training.

\renewcommand\px{\mathbf{x}}
\renewcommand{\calL}{\mathcal{L}}

Supervised training is performed for the pre-training with ground-truth annotations of optical flow, disparity, and their associated occlusions. The loss function is defined as
\begin{align}
\calL_{sp} = \left(\calL_F + \calL_{O_F}\right) +  0.25\times\left(\calL_D + \calL_{O_D}\right).
\end{align}
For Monkaa and Driving, since only optical flow and disparity annotations are available, we only set $\calL_{O_D}$ and $\calL_{O_F}$ to 0 for training data sampled from Monkaa and Driving.

During training, we use color jittering, including randomly changing gamma value, changing brightness, changing contrast, and adding Gaussian noise, for both optical flow and disparity training. Additionally, we use random crops and vertical flips for stereo training images. The crop size is $256\times 512$. For optical flow training images, we perform extensive data augmentations  including random crop, translation, rotation, zooming, squeezing, and horizontal and vertical flip, where the crop size is $384\times 640$.
The network is trained for 100 epochs with a batch size of 8 using the Adam optimizer~\cite{kingma14adam}. We use synchronized Batch Normalization~\cite{xiao18unified} to ensure there are enough training samples for estimating Batch Normalization layers' statistics when using multiple GPUs. The initial learning rate is 0.001 and decreased by factor of 10 after 70 epochs.

\noindent\textbf{Sintel.} We fine-tune the pre-trained model on Sintel. Sintel training data provides optical flow, disparity, and their corresponding occlusion annotations. We therefore use the same loss function as used for the pre-training.

During training, we apply the same color jittering used for pre-training. Similarly we use random crops and vertical flips for stereo training images with crop size of $384\times 768$. For optical flow training images, we perform extensive data augmentations as well including random crop, translation, rotation, zooming, squeezing, and horizontal and vertical flip, where the crop size is $384\times 768$.

Synchronized Batch Normalization is used with batch size of 8. The model is first trained for 500 epochs using the Adam optimizer with an initial learning rate of 0.0005, which is decreased by factor of 2 after every 100 epochs. The weight decay is 0.0004. After 500-epoch training is finished, we keep fine-tuning the model for another 500 epochs using Adam with an initial learning rate of 0.0002, which is decreased by factor of 2 after every 100 epochs. The weight decay remains~0.0004.

\noindent\textbf{KITTI.} On KITTI (including KITTI2012 and KITTI2015), we use both supervised loss and semi-supervised loss. The final loss is defined as
\begin{align}
\calL = \underbrace{\calL_F + \calL_D}_\text{supervised loss} + \underbrace{\alpha_O\big(\calL_{O_{Fd}} + \calL_{O_{Dd}}\big) + \alpha_{S_d}\calL_{S_d}}_\text{distillation loss} + \underbrace{\alpha_{PC}\calL_{PC} + \alpha_{SC}\calL_{SC} + \calL_{SS} + \calL_{REG}}_\text{self-supervised loss},
\end{align}
where $\calL_{O_{Fd}}$ and $\calL_{O_{Dd}}$ are distillation loss for optical flow occlusion and disparity occlusion, respectively. They are defined as \texttt{smooth\_L1} loss between the pseudo ground-truth (\ie, estimations from a model pre-trained on synthetic SceneFlow dataset) and estimations from the model being trained. On the validation set, we empirically found $\alpha_O=0.05, \alpha_{S_d}=1, \alpha_{PC}=0.5, \alpha_{SC}=0.5$ work well. For the SSIM loss, we use $\gamma_D=0.005\times C_H\times C_W$ and $\gamma_{F}=0.01\times C_H\times C_W$\footnote{In our definition of SSIM loss, the function $SS(\cdot, \cdot)$ gives a single scalar value.}, where $C_H$ and $C_W$ are crop height and width, respectively. For the regularization term, we empirically set $\beta_F=\beta_D=0.5$.

During training, we use similar color jittering used in pre-training but with a probability of 0.5. Similarly we use random crops and vertical flips for stereo training images with crop size of $320\times 768$. For optical flow training images, we perform extensive data augmentations as well including random crop, translation, rotation, zooming, squeezing, and horizontal and vertical flip, where the crop size is $320\times 768$.

Synchronized Batch Normalization is used with batch size of 8. The model is fine-tuned for 1,500 epochs using the Adam optimizer with an initial learning rate of 0.001, which is decreased by factor of 2 at epochs of 400, 800, 1,000, 1200, and 1,400. The weight decay is 0.0004. Another round of fine-tuning is followed with an initial learning rate of 0.0002, which is decreased by factor of 2 at epochs of 400, 600, 800, and 900.

\section{Rigidity-based Warped Disparity Refinement for Scene Flow Estimation}

\noindent \textbf{Determine rigidity area.} Given the estimated semantic segmentation labels of the first left frame $\mathbf{S}^{1,l}$, we select pixels as static rigid regions by removing pixels which have a semantic label of vehicle, pedestrian, cyclist, or sky. This step gives a conservative selection of static regions with points not at infinity. The output is a binary mask $\mathbf{B}$ with the label 1 indicating static rigid region. Since the semantic segmentation can be inaccurate at object boundary, we further perform an erosion operation with a size of 10 on the static rigid region mask $\mathbf{B}$.

\noindent \textbf{Estimate rigid flow induced by camera motion.} Given the estimated flow $\mathbf{F}^{1}$ and disparity $\mathbf{D}^{1}$ of the left frame, we calculate the ego-motion flow induced by the rigid camera motion by minimizing the weighted errors between predicted rigid flow $\mathbf{F}^{1}_{\mathbf{R}}$ and optical flow $\mathbf{F}^{1}$ in the background region pixels $\mathbf{x} \in \mathbf{R}^2$:
\begin{align}
    \argmin_{\xi} &  \mathbf{r}^{T}(\xi;\mathbf{x}) \mathbf{W} \mathbf{r}(\xi;\mathbf{x}) \label{eq:rigid_objective} \\
    \mathbf{r}(\xi;\mathbf{x}) &= \mathbf{F}^{1}(\mathbf{x}) - \mathbf{F}^1_{\mathbf{R}}(\xi;\mathbf{x}) \\
    \mathbf{F}^1_{\mathbf{R}}(\xi;\mathbf{x}) &= \mathcal{W}(\xi; \mathbf{x}, \mathbf{D}^{1}(\mathbf{x}) ) - \mathbf{x}
\end{align}

\noindent where $\mathbf{x} \in \mathbf{R}^{2}$ denotes the pixels in 2D image space which are within the rigid areas $\mathbf{B}$. $\mathcal{W}(\xi;\mathbf{x}, \mathbf{D}^{1})$ is the warping function which transforms the pixels $\mathbf{x}$ and its corresponding disparity $\mathbf{D}^{1}(\mathbf{x})$ with an estimated transform $\xi \in \mathbf{SE}(3)$. $\mathbf{W}$ is a diagonal weight matrix that depends on residuals using Huber weight function.

We solve equation $\ref{eq:rigid_objective}$ as an iteratively reweighted least-square problem using Gauss-Newton update:
\begin{align}
    \delta \xi &= (\mathbf{J}^{T} \mathbf{W} \mathbf{J})^{-1} \mathbf{J}^{T} \mathbf{W} \mathbf{r}  \\
    \xi &= \xi \circ \delta \xi
\end{align}

\noindent where $\circ$ indicates the right composition of $\xi \in \mathbf{SE}(3)$. $\mathbf{J}$ is the Jacobian matrix of $\partial \mathbf{F}^{1}_{\mathbf{R}}(\xi) / \partial \xi$.

Suppose $\mathbf{K}$ is the intrinsic matrix for a pin-hole camera without distortion, which can be parameterized as $(f_x, f_y, c_x, c_y)$ with $f_x, f_y$ as its focal length and $c_x, c_y$ as its offset along the two axes. The baseline of the stereo pair is $b$. We define the 3D point $\mathbf{p}=(p_x, p_y, p_z)$ as $\mathbf{p}=(f_x b/\mathbf{D}^{1}(\mathbf{x}))\mathbf{K}^{-1}\mathbf{x}$. Through chain-rule, we can derive the analytical form of the Jacobian matrix $\mathbf{J}$. To simplify the computation, we use the inverse depth parameterization $\mathbf{p}=(p_u/p_d, p_v/p_d, 1/p_d)$ in which $\mathbf{x}=(p_u, p_v) \in \mathbf{R}^2$ is the pixel of coordinate of $\mathbf{x}$ and $p_d$ is the inverse depth as $p_d = \mathbf{D}^{1}(\mathbf{x})/(f_x b)$. Thus, we obtain the Jacobian matrix at a pixel $\mathbf{x}$ as:

\begin{align}
\begin{bmatrix}
-p_u p_v f_x & (1+p_u^2)f_x & -p_v f_x & p_d f_x & 0 & -p_u p_d f_x \\
-(1+p_v^2)f_y & p_u p_v f_y & p_u f_y & 0 & p_d f_y & -p_u p_d f_y
\end{bmatrix}
\end{align}

We perform the Gauss-Newton update if the absolute residual error is bigger than $10^{-6}$ with a maximum of 20 iterations. All operations are implemented in Pytorch and executed in GPU. The running time of the total optimization varies between 0.03s and 0.2s, according to the number of iterations. In average, the optimization step takes 0.1s for KITTI image of resolution 375x1242.

The final optical flow $\mathbf{F}$ is an element-wise linear composition of $\mathbf{F}^{1}$ and $\mathbf{F}^1_{\mathbf{R}}$ as:

\begin{equation}
    \mathbf{F} = (\mathbf{1} - \mathbf{B})\otimes \mathbf{F}^{1} + \mathbf{B}\otimes \mathbf{F}^1_{\mathbf{R}}
\end{equation}

\noindent where $\otimes$ indicates element-wise multiplications.

\noindent \textbf{Estimate warped second frame rigid disparity.} Given the estimated optimal $\xi^{\star}$, we define the disparity $\mathbf{D}^{1\rightarrow2}_{\mathcal{W},\mathbf{R}}$ of the second frame warped from the first frame following the optimal rigid transform $\xi^{\star}$:

\begin{equation}
\mathbf{D}^{1\rightarrow2}_{\mathcal{W},\mathbf{R}} = \mathcal{W}_{\mathbf{D}}^{1\rightarrow2}(\xi^{\star}; \mathbf{D}^{1})
\end{equation}

\noindent where $\mathcal{W}_{\mathbf{D}}^{1\rightarrow2}(\cdot)$ defines the disparity channel output from the warping function $\mathcal{W}(\cdot)$ through a forward warping. Given the forward optical flow $\mathbf{F}^{1}$, the warped disparity of the second frame can be computed through an inverse warping $\mathcal{W}_{\mathbf{D}}^{2\rightarrow1}$ as:

\begin{equation}
\mathbf{D}^{2 \rightarrow 1}_{\mathcal{W}} = \mathcal{W}_{\mathbf{D}}^{2\rightarrow1}(\mathbf{F}^{1}, \mathbf{D}^2)
\end{equation}

We find that the disparity through forward warping $\mathbf{D}^{1\rightarrow2}_{\mathcal{W},\mathbf{R}}$ gives more accurate disparity in static region and can better handle occlusions. The final warped disparity $\mathbf{D}^2_{\mathcal{W}}$ is a element-wise linear composition of $\mathbf{D}^{2 \rightarrow 1}_{\mathcal{W}}$ and $\mathbf{D}^{1\rightarrow2}_{\mathcal{W},\mathbf{R}}$ as:

\begin{equation}
    \mathbf{D}^2_{\mathcal{W}} = (\mathbf{1}-\mathbf{B}) \otimes \mathbf{D}^{2 \rightarrow 1}_{\mathcal{W}} + \mathbf{B} \otimes
    \mathbf{D}^{1\rightarrow2}_{\mathcal{W},\mathbf{R}}
\end{equation}

Note that both warping function cannot deal with out-of-boundary pixels due to two-view occlusion. This can be resolved by the additional refinement network detailed in the following section.

\begin{table}[h!]
\centering
\caption{Definition of our shared encoder. $H$ and $W$ denote the height and width of the input images. $[\cdot]$ indicates a residual block~\cite{he16deep}. We use convolution with stride of 2 to perform downsampling. The first downsampling is performed in the layer conv1\_1.}
\label{tab:def_encoder}
\begin{tabular}{c|c|c}
\toprule
layer name & output size & layer setting \\
\midrule
input & $H\times W\times 3$ & - \\
\midrule
conv1\_1 & \multirow{3}{*}{$\frac{H}{2}\times \frac{W}{2}\times 32$} & $3\times3$, 32 \\
conv1\_2 &  & $3\times3$, 32 \\
conv1\_3 &  & $3\times3$, 32 \\
\midrule
conv2 & $\frac{H}{4}\times \frac{W}{4}\times 32$ & $\left[ \begin{array}{l} 3\times 3, 32 \\ 3\times 3, 32 \end{array} \right] \times 3$ \\
\midrule
conv3 & $\frac{H}{8}\times \frac{W}{8}\times 64$ & $\left[\begin{array}{l} 3\times 3, 64 \\ 3\times 3, 64 \end{array}\right]\times 16$ \\
\midrule
conv4 & $\frac{H}{16}\times \frac{W}{16}\times 128$ & $\left[\begin{array}{l} 3\times 3, 128 \\ 3\times 3, 128 \end{array}\right] \times 3$ \\
\midrule
conv5 & $\frac{H}{32} \times \frac{W}{32}\times 128$ & $\left[\begin{array}{l} 3\times 3, 128 \\ 3\times 3, 128 \end{array}\right]\times 3$ \\
\bottomrule
\end{tabular}
\end{table}

\section{Details of Network Architecture}

\begin{itemize}
    \item Table~\ref{tab:def_encoder} provides the detailed network architecture of our shared encoder, which is a  ResNet-like~\cite{he16deep} architecture.
\item Table~\ref{tab:def_ppm} provides details of the pyramid pooling module (PPM), which aggregates multi-scale feature maps to enhance disparity estimation and semantic segmentation.
\item The Hourglass module used for disparity estimation refinement is illustrated in Table~\ref{tab:def_hourglass}. The input is a concatenation of upsampled disparity (by a factor of 2), the feature map of the first image (128-dimensional), and the warped feature map of the second image (128-dimensional). The output is a residual disparity estimation that is added to the twice upsampled disparity.
\item The refinement network for warped disparity, which is used for scene flow estimation, can be found in Table~\ref{tab:def_refinement}. The input is a concatenation of $\mv{I}^{1,l}$, $\mv{O}_F^{1,l}$, $\mv{D}^{1,l}$, and $g(\mv{D}^{2,l}, \mv{F}^{1,l})$, with 6 (=3+1+1+1) channels in total. The network consists of an encoder, a decoder, and skip connections between them. It contains 9.1M parameters in total and takes 0.01s for inference for a KITTI image with resolution of $375\times 1242$. We perform supervision for intermediate layers of the decoder, in a manner similar to optical flow and disparity estimations.
\end{itemize}

\begin{table}[t]
\centering
\caption{Definition of our PPM head, where branch1, branch2, branch3, and branch4 are all parallel branches on top of the conv5 layer in the encoder.}
\label{tab:def_ppm}
\begin{tabular}{c|c|c}
\toprule
layer name & output size & layer setting \\
\midrule
\multirow{3}{*}{branch1} & \multirow{3}{*}{$\frac{H}{32} \times \frac{W}{32}\times 128$} & $1\times 1$ adaptive avg. pool \\
& & $1\times 1, 128$ \\
& & bilinear interpolation \\
\midrule
\multirow{3}{*}{branch2} & \multirow{3}{*}{$\frac{H}{32} \times \frac{W}{32}\times 128$} & $2\times 2$ adaptive avg. pool \\
& & $1\times 1, 128$ \\
& & bilinear interpolation \\
\midrule
\multirow{3}{*}{branch3} & \multirow{3}{*}{$\frac{H}{32} \times \frac{W}{32}\times 128$} & $3\times 3$ adaptive avg. pool \\
& & $1\times 1, 128$ \\
& & bilinear interpolation \\
\midrule
\multirow{3}{*}{branch4} & \multirow{3}{*}{$\frac{H}{32} \times \frac{W}{32}\times 128$} & $6\times 6$ adaptive avg. pool \\
& & $1\times 1, 128$ \\
& & bilinear interpolation \\
\midrule
\multirow{3}{*}{fusion} & \multirow{3}{*}{$\frac{H}{32} \times \frac{W}{32}\times 128$} & concat of conv5, branch1 \\
& & branch2, branch3, and branch4 \\
& & $3\times 3, 128$ \\
\bottomrule
\end{tabular}
\end{table}

\begin{table}[]
\centering
\small
\caption{Definition of our disparity Hourglass refinement model.}
\label{tab:def_hourglass}
\begin{tabular}{c|c|c}
\toprule
layer name & output size & layer setting \\
\midrule
input & $\frac{H}{2} \times \frac{W}{2}\times 257$ & - \\
\midrule
conv1 & $\frac{H}{4} \times \frac{W}{4}\times 514$ & $3\times 3, 514$ \\
\midrule
conv2 & $\frac{H}{8} \times \frac{W}{8}\times 514$ & $3\times 3, 514$ \\
\midrule
conv3 & $\frac{H}{8} \times \frac{W}{8}\times 514$ & $3\times 3, 514$ \\
\midrule
\multirow{2}{*}{conv4} & \multirow{2}{*}{$\frac{H}{4} \times \frac{W}{4}\times 514$} & bilinear interpolation \\
& & $3\times 3, 514$ \\
\midrule
\multirow{2}{*}{conv5} & \multirow{2}{*}{$\frac{H}{2} \times \frac{W}{2}\times 257$} & bilinear interpolation \\
& & $3\times 3, 257$ \\
\midrule
output & $\frac{H}{2}\times \frac{W}{2}\times 1$ & $3\times 3$, 1 \\
\bottomrule
\end{tabular}
\end{table}

\begin{table}[]
\centering
\small
\caption{Definition of our warped disparity refinement model. output5 is on top of decoder\_layer\_5. output4, output3, and output2 are computed similarly. output1 is on top of decoder\_layer1\_2. We compute loss for all five output during training and sum them up as the final loss. During inference, we only compute output1.}
\label{tab:def_refinement}
\begin{tabular}{c|c|c}
\toprule
layer name & output size & layer setting \\
\midrule
input & $H \times W\times 6$ & - \\
\midrule
encoder\_layer1 & $H \times W\times 32$ & $3\times 3, 32$ \\
\midrule
\multirow{2}{*}{encoder\_layer2} & \multirow{2}{*}{$\frac{H}{2} \times \frac{W}{2}\times 32$} & $2\times 2$ avg. pool, stride of 2 \\
& & $3\times 3, 32$ \\
\midrule
\multirow{2}{*}{encoder\_layer3} & \multirow{2}{*}{$\frac{H}{4} \times \frac{W}{4}\times 32$} & $2\times 2$ avg. pool, stride of 2 \\
& & $3\times 3, 32$ \\
\midrule
\multirow{2}{*}{encoder\_layer4} & \multirow{2}{*}{$\frac{H}{8} \times \frac{W}{8}\times 32$} & $2\times 2$ avg. pool, stride of 2 \\
& & $3\times 3, 32$ \\
\midrule
\multirow{2}{*}{encoder\_layer5} & \multirow{2}{*}{$\frac{H}{16} \times \frac{W}{16}\times 32$} & $2\times 2$ avg. pool, stride of 2 \\
& & $3\times 3, 32$ \\
\midrule
\multirow{2}{*}{bottleneck} & \multirow{2}{*}{$\frac{H}{32} \times \frac{W}{32}\times 32$} & $2\times 2$ avg. pool, stride of 2 \\
& & $3\times 3, 32$ \\
\midrule
\multirow{2}{*}{decoder\_layer5} & \multirow{2}{*}{$\frac{H}{16} \times \frac{W}{16}\times 512$} & $3\times 3, 512$ \\
& & $2\times$ bilinear interpolation \\
\midrule
\multirow{2}{*}{decoder\_layer4} & \multirow{2}{*}{$\frac{H}{8} \times \frac{W}{8}\times 256$} & concat. with encoder\_layer\_5 \\
& & $3\times 3, 256$ \\
& & $2\times$ bilinear interpolation \\
\midrule
\multirow{2}{*}{decoder\_layer3} & \multirow{2}{*}{$\frac{H}{4} \times \frac{W}{4}\times 128$} & concat. with encoder\_layer\_4 \\
& & $3\times 3, 128$ \\
& & $2\times$ bilinear interpolation \\
\midrule
\multirow{2}{*}{decoder\_layer2} & \multirow{2}{*}{$\frac{H}{2} \times \frac{W}{2}\times 64$} & concat. with encoder\_layer\_3 \\
& & $3\times 3, 64$ \\
& & $2\times$ bilinear interpolation \\
\midrule
\multirow{3}{*}{decoder\_layer1\_1} & \multirow{3}{*}{$H \times W\times 32$} & concat. with encoder\_layer\_2 \\
& & $3\times 3, 32$ \\
& & $2\times$ bilinear interpolation \\
\midrule
\multirow{2}{*}{decoder\_layer1\_2} & \multirow{2}{*}{$H \times W\times 32$} & concat. with encoder\_layer\_1 \\
& & $3\times 3, 32$ \\
\midrule
output5 & $\frac{H}{16}\times \frac{W}{16}\times 1$ & $3\times 3, 1$ \\
\midrule
output4 & $\frac{H}{8}\times \frac{W}{8}\times 1$ & $3\times 3, 1$ \\
\midrule
output3 & $\frac{H}{4}\times \frac{W}{4}\times 1$ & $3\times 3, 1$ \\
\midrule
output2 & $\frac{H}{2}\times \frac{W}{2}\times 1$ & $3\times 3, 1$ \\
\midrule
output1 & $H\times W\times 1$ & $3\times 3, 1$ \\
\bottomrule
\end{tabular}
\end{table}

\section{More Ablation Studies}

\begin{table}[]
\caption{Ablation study of design choices for disparity.}
\label{tab:ablation_net_design}
\renewcommand{\tabcolsep}{8pt}
\centering
\footnotesize
\begin{tabular}{cccc|ccc}
\toprule
\rotatebox{0}{{\scriptsize 5 layers+BN}} & \rotatebox{0}{{\scriptsize ResNet encoder}} & \rotatebox{0}{{\scriptsize PPM}} & \rotatebox{0}{{\scriptsize hourglass}} &
\rotatebox{0}{\#params} & \rotatebox{0}{{\scriptsize FlyThings3D (EPE)}} & \rotatebox{0}{\scriptsize{KITTI2015} (EPE)} \\
\midrule
 & & & & 7.1M & 2.10 & 1.01 \\
\midrule
\checkmark & & & & 3.6M & 1.61 & 0.85 \\
\checkmark & \checkmark & & & 6.9M & 1.40 & 0.77 \\
\checkmark & \checkmark & \checkmark & & 7.7M & 1.32 & 0.77 \\
\checkmark & \checkmark & \checkmark & \checkmark & 8.3M & 1.15 & 0.71 \\
\bottomrule
\end{tabular}
\end{table}

\noindent\textbf{New network design.} As shown in Table~\ref{tab:ablation_net_design}, we study the effectiveness of new network designs for disparity estimation. The baseline model has exactly the same architecture as PWC-Net~\cite{sun2018pwc} for optical flow estimation, except we construct  a 1D cost volume for disparity estimation. It is a compact model with 7.1M parameters. However, most of the parameters concentrate in the decoder due to DenseNet blocks. By removing the last pyramid in both encoder and decoder and adding Batch Normalization layers, we obtain significant improvement in disparity while halving the parameters. By replacing the original encoder consisting of plain CNN layers with deeper residual blocks, we obtain further improvements and yet still have fewer parameters. Adding PPM and hourglass refinement keeps improving the accuracy. Our final model has slightly more parameters than the baseline, but the performance on both synthetic and real-world benchmark datasets increases substantially.

\noindent\textbf{Shared encoder.} We report optical flow and disparity errors on Sintel, KITTI 2012, and KITTI 2015 using both separate and shared encoders in Table~\ref{tab:effect_shared_encoder}, where a model trained on synthetic SceneFlow dataset is used. As we can see, a shared encoder leads to better EPE metrics on both KITTI 2012 and KITTI 2015.

\section{Visual Results of Optical Flow and Disparity Estimations}
\begin{figure*}[t]
\centering
\renewcommand{\tabcolsep}{0.25mm}
\newcommand{\FigWidth}{0.33\linewidth}
\renewcommand{\arraystretch}{0.6}
\begin{tabular}{ccc}
\includegraphics[width=\FigWidth]{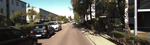}&
\includegraphics[width=\FigWidth]{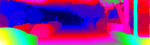}&
\includegraphics[width=\FigWidth]{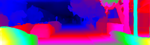}\\
\includegraphics[width=\FigWidth]{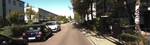}&
\includegraphics[width=\FigWidth]{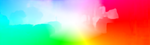}&
\includegraphics[width=\FigWidth]{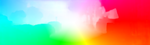}\\
\includegraphics[width=\FigWidth]{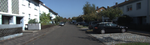}&
\includegraphics[width=\FigWidth]{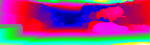}&
\includegraphics[width=\FigWidth]{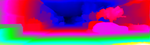}\\
\includegraphics[width=\FigWidth]{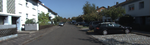}&
\includegraphics[width=\FigWidth]{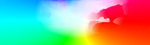}&
\includegraphics[width=\FigWidth]{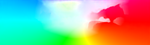}\\
\includegraphics[width=\FigWidth]{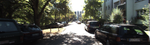}&
\includegraphics[width=\FigWidth]{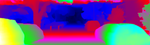}&
\includegraphics[width=\FigWidth]{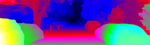}\\
\includegraphics[width=\FigWidth]{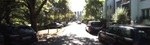}&
\includegraphics[width=\FigWidth]{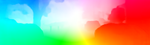}&
\includegraphics[width=\FigWidth]{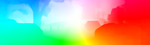}\\
\includegraphics[width=\FigWidth]{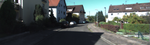}&
\includegraphics[width=\FigWidth]{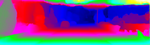}&
\includegraphics[width=\FigWidth]{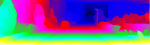}\\
\includegraphics[width=\FigWidth]{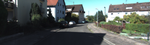}&
\includegraphics[width=\FigWidth]{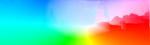}&
\includegraphics[width=\FigWidth]{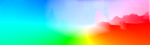}\\
\includegraphics[width=\FigWidth]{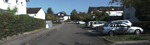}&
\includegraphics[width=\FigWidth]{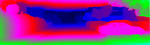}&
\includegraphics[width=\FigWidth]{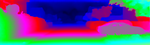}\\
\includegraphics[width=\FigWidth]{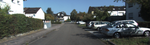}&
\includegraphics[width=\FigWidth]{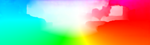}&
\includegraphics[width=\FigWidth]{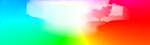}\\
\includegraphics[width=\FigWidth]{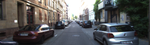}&
\includegraphics[width=\FigWidth]{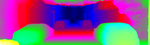}&
\includegraphics[width=\FigWidth]{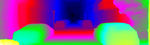}\\
\includegraphics[width=\FigWidth]{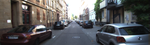}&
\includegraphics[width=\FigWidth]{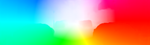}&
\includegraphics[width=\FigWidth]{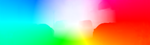}\\
(a) input images & (b) supervised loss & (c) full loss \\
\end{tabular}
\caption{Visual results on the test set of KITTI 2012. We show two consecutive video frames in the first column. In the second and third columns, we show disparity in every first row and optical flow in the other one. Best viewed in color.}
\label{fig:kitti2012_part1}
\end{figure*}

\begin{figure*}[t]
\centering
\renewcommand{\tabcolsep}{0.25mm}
\newcommand{\FigWidth}{0.33\linewidth}
\renewcommand{\arraystretch}{0.6}
\begin{tabular}{ccc}
\includegraphics[width=\FigWidth]{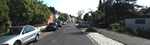}&
\includegraphics[width=\FigWidth]{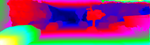}&
\includegraphics[width=\FigWidth]{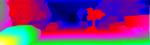}\\
\includegraphics[width=\FigWidth]{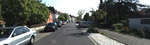}&
\includegraphics[width=\FigWidth]{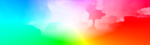}&
\includegraphics[width=\FigWidth]{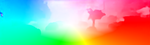}\\
\includegraphics[width=\FigWidth]{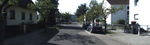}&
\includegraphics[width=\FigWidth]{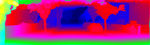}&
\includegraphics[width=\FigWidth]{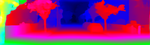}\\
\includegraphics[width=\FigWidth]{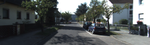}&
\includegraphics[width=\FigWidth]{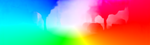}&
\includegraphics[width=\FigWidth]{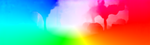}\\
\includegraphics[width=\FigWidth]{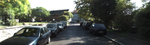}&
\includegraphics[width=\FigWidth]{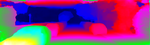}&
\includegraphics[width=\FigWidth]{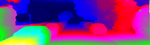}\\
\includegraphics[width=\FigWidth]{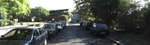}&
\includegraphics[width=\FigWidth]{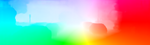}&
\includegraphics[width=\FigWidth]{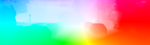}\\
\includegraphics[width=\FigWidth]{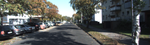}&
\includegraphics[width=\FigWidth]{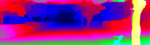}&
\includegraphics[width=\FigWidth]{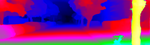}\\
\includegraphics[width=\FigWidth]{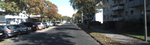}&
\includegraphics[width=\FigWidth]{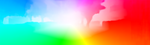}&
\includegraphics[width=\FigWidth]{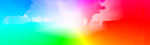}\\
\includegraphics[width=\FigWidth]{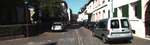}&
\includegraphics[width=\FigWidth]{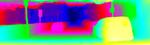}&
\includegraphics[width=\FigWidth]{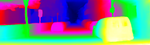}\\
\includegraphics[width=\FigWidth]{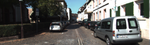}&
\includegraphics[width=\FigWidth]{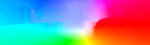}&
\includegraphics[width=\FigWidth]{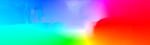}\\
\includegraphics[width=\FigWidth]{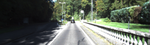}&
\includegraphics[width=\FigWidth]{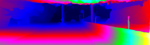}&
\includegraphics[width=\FigWidth]{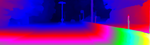}\\
\includegraphics[width=\FigWidth]{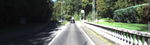}&
\includegraphics[width=\FigWidth]{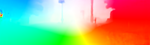}&
\includegraphics[width=\FigWidth]{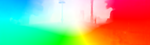}\\
(a) input images & (b) supervised loss & (c) full loss \\
\end{tabular}
\caption{Visual results on the test set of KITTI 2012. We show two consecutive video frames in the first column. In the second and third columns, we show disparity in every first row and optical flow in the other one. Best viewed in color.}
\label{fig:kitti2012_part2}
\end{figure*}

\begin{figure*}[t]
\centering
\renewcommand{\tabcolsep}{0.25mm}
\newcommand{\FigWidth}{0.33\linewidth}
\renewcommand{\arraystretch}{0.6}
\begin{tabular}{ccc}
\includegraphics[width=\FigWidth]{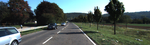}&
\includegraphics[width=\FigWidth]{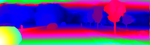}&
\includegraphics[width=\FigWidth]{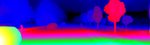}\\
\includegraphics[width=\FigWidth]{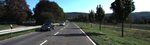}&
\includegraphics[width=\FigWidth]{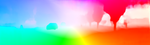}&
\includegraphics[width=\FigWidth]{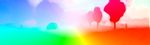}\\
\includegraphics[width=\FigWidth]{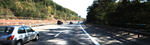}&
\includegraphics[width=\FigWidth]{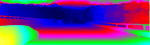}&
\includegraphics[width=\FigWidth]{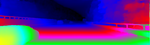}\\
\includegraphics[width=\FigWidth]{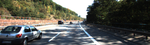}&
\includegraphics[width=\FigWidth]{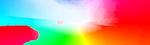}&
\includegraphics[width=\FigWidth]{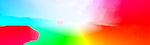}\\
\includegraphics[width=\FigWidth]{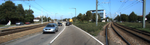}&
\includegraphics[width=\FigWidth]{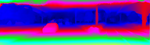}&
\includegraphics[width=\FigWidth]{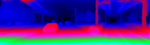}\\
\includegraphics[width=\FigWidth]{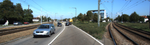}&
\includegraphics[width=\FigWidth]{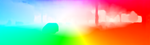}&
\includegraphics[width=\FigWidth]{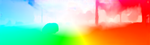}\\
\includegraphics[width=\FigWidth]{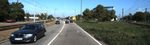}&
\includegraphics[width=\FigWidth]{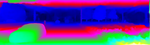}&
\includegraphics[width=\FigWidth]{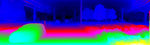}\\
\includegraphics[width=\FigWidth]{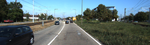}&
\includegraphics[width=\FigWidth]{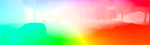}&
\includegraphics[width=\FigWidth]{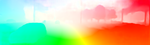}\\
\includegraphics[width=\FigWidth]{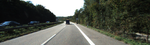}&
\includegraphics[width=\FigWidth]{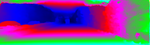}&
\includegraphics[width=\FigWidth]{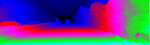}\\
\includegraphics[width=\FigWidth]{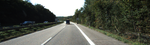}&
\includegraphics[width=\FigWidth]{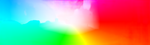}&
\includegraphics[width=\FigWidth]{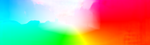}\\
\includegraphics[width=\FigWidth]{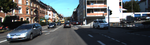}&
\includegraphics[width=\FigWidth]{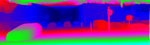}&
\includegraphics[width=\FigWidth]{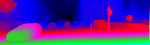}\\
\includegraphics[width=\FigWidth]{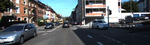}&
\includegraphics[width=\FigWidth]{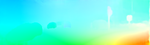}&
\includegraphics[width=\FigWidth]{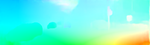}\\
(a) input images & (b) supervised loss & (c) full loss \\
\end{tabular}
\caption{Visual results on the test set of KITTI 2015. We show two consecutive video frames in the first column. In the second and third columns, we show disparity in every first row and optical flow in the other one. Best viewed in color.}
\label{fig:kitti2015_part1}
\end{figure*}

\begin{figure*}[t]
\centering
\renewcommand{\tabcolsep}{0.25mm}
\newcommand{\FigWidth}{0.33\linewidth}
\renewcommand{\arraystretch}{0.6}
\begin{tabular}{ccc}
\includegraphics[width=\FigWidth]{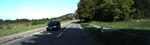}&
\includegraphics[width=\FigWidth]{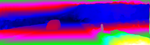}&
\includegraphics[width=\FigWidth]{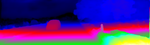}\\
\includegraphics[width=\FigWidth]{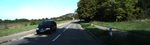}&
\includegraphics[width=\FigWidth]{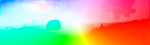}&
\includegraphics[width=\FigWidth]{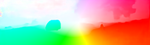}\\
\includegraphics[width=\FigWidth]{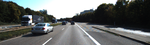}&
\includegraphics[width=\FigWidth]{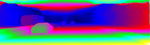}&
\includegraphics[width=\FigWidth]{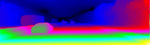}\\
\includegraphics[width=\FigWidth]{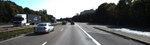}&
\includegraphics[width=\FigWidth]{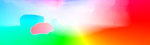}&
\includegraphics[width=\FigWidth]{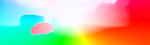}\\
\includegraphics[width=\FigWidth]{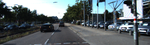}&
\includegraphics[width=\FigWidth]{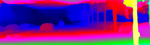}&
\includegraphics[width=\FigWidth]{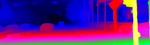}\\
\includegraphics[width=\FigWidth]{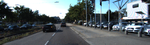}&
\includegraphics[width=\FigWidth]{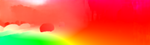}&
\includegraphics[width=\FigWidth]{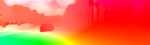}\\
\includegraphics[width=\FigWidth]{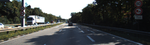}&
\includegraphics[width=\FigWidth]{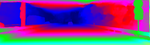}&
\includegraphics[width=\FigWidth]{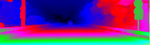}\\
\includegraphics[width=\FigWidth]{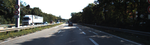}&
\includegraphics[width=\FigWidth]{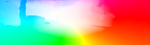}&
\includegraphics[width=\FigWidth]{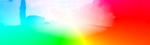}\\
\includegraphics[width=\FigWidth]{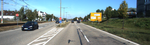}&
\includegraphics[width=\FigWidth]{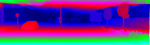}&
\includegraphics[width=\FigWidth]{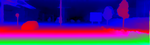}\\
\includegraphics[width=\FigWidth]{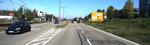}&
\includegraphics[width=\FigWidth]{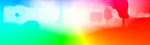}&
\includegraphics[width=\FigWidth]{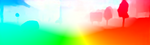}\\
\includegraphics[width=\FigWidth]{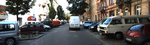}&
\includegraphics[width=\FigWidth]{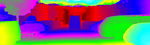}&
\includegraphics[width=\FigWidth]{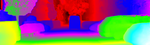}\\
\includegraphics[width=\FigWidth]{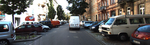}&
\includegraphics[width=\FigWidth]{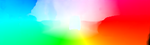}&
\includegraphics[width=\FigWidth]{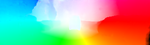}\\
(a) input images & (b) supervised loss & (c) full loss \\
\end{tabular}
\caption{Visual results on the test set of KITTI 2015. We show two consecutive video frames in the first column. In the second and third columns, we show disparity in every first row and optical flow in the other one. Best viewed in color.}
\label{fig:kitti2015_part2}
\end{figure*}

We provide more visual results of optical flow and disparity estimations of the test set in Fig.~\ref{fig:kitti2012_part1} and Fig.~\ref{fig:kitti2012_part2} for KITTI 2012 and in Fig.~\ref{fig:kitti2015_part1} and Fig.~\ref{fig:kitti2015_part2} for KITTI 2015.

We can clearly see our full model (supervised loss plus semi-supervised loss) produces visually better results on both KITTI 2012 and KITTI 2015.

\begin{table}[t!]
\centering
\caption{Effectiveness of the shared encoder for optical flow and disparity estimations.}
\label{tab:effect_shared_encoder}
\begin{tabular}{cc|cc|cc|cc}
\toprule
& & \multicolumn{2}{c|}{Sintel (EPE)} & \multicolumn{2}{c|}{KITTI 2012} & \multicolumn{2}{c}{KITTI 2015} \\
\cmidrule(lr){3-4}\cmidrule(lr){5-6}\cmidrule(lr){7-8}
& & clean & final & EPE & D1/F1-occ & EPE & D1/F1-occ \\
\midrule
\multirow{2}{*}{optical flow} & separate encoder & 1.97 & 3.34 & 2.63 & 11.72\% & 6.37 & 21.15\% \\
 & shared encoder & 1.91 & 3.78 & 2.55 & 12.56\% & 6.23 & 23.29\% \\
\midrule
\multirow{2}{*}{disparity} & separate encoder & 1.56 & 2.99 & 1.09 & 6.17\% & 1.26 & 6.62\% \\
 & shared encoder & 1.70 & 3.20 & 1.04 & 5.42\% & 1.22 & 6.38\% \\
\bottomrule
\end{tabular}
\end{table}

{
\section*{Acknowledgement}
Huaizu Jiang and Erik Learned-Miller acknowledge support from AFRL and DARPA (\#FA8750-
18-2-0126) and the MassTech Collaborative grant for funding the UMass GPU cluster.
The U.S. Gov. is authorized to reproduce and distribute reprints for Gov.
purposes notwithstanding any copyright notation thereon. The views and
conclusions contained herein are those of the authors and should not be interpreted as necessarily representing the official policies or endorsements,
either expressed or implied, of the AFRL and DARPA or the U.S. Gov.
}

\end{appendices}

\end{document}